\documentclass{INTERSPEECH2023}

\usepackage{xcolor}
\usepackage{multirow}
\usepackage[flushleft]{threeparttable}

\interspeechcameraready

\title{On-Device Constrained Self-Supervised Speech Representation Learning for Keyword Spotting via Knowledge Distillation}
\name{
Gene-Ping Yang$^{1\dagger}$\thanks{$^{\dagger}$Work done at Amazon}, Yue Gu$^2$, Qingming Tang$^2$, Dongsu Du$^2$, Yuzong Liu$^{3\dagger}$
}
\address{
  $^1$Centre for Speech Technology Research, University of Edinburgh\\
  $^2$Alexa Perceptual Technologies, Amazon $\quad$ $^3$Zoom Video Communications, Inc.
  }
\email{geneping.yang@ed.ac.uk, \{yguam,qmtang,dudongsu\}@amazon.com, yuzong.liu@zoom.us}

\begin{document}

\maketitle
 
\begin{abstract}
\vspace{-0.5em}
Large self-supervised models are effective feature extractors, but their application is challenging under on-device budget constraints and biased dataset collection, especially in keyword spotting. 
To address this, we proposed a knowledge distillation-based self-supervised speech representation learning (S3RL) architecture for on-device keyword spotting. 
Our approach used a teacher-student framework to transfer knowledge from a larger, more complex model to a smaller, light-weight model using dual-view cross-correlation distillation and the teacher's codebook as learning objectives. 
We evaluated our model's performance on an Alexa keyword spotting detection task using a 16.6k-hour in-house dataset. 
Our technique showed exceptional performance in normal and noisy conditions, demonstrating the efficacy of knowledge distillation methods in constructing self-supervised models for keyword spotting tasks while working within on-device resource constraints.
\end{abstract}

\section{Introduction}
\vspace{-0.0em}

Self-supervised methods have proven highly effective as general feature extractors for various speech-related tasks \cite{mohamed2022self, yang21c_interspeech}.
Unlike supervised methods, the self-supervised models are trained to do autoregressive prediction \cite{CHTG2019, ling2020decoar, yang2022}, contrastive learning \cite{VLV2018, BZMA2020}, mask reconstruction \cite{9053541,JLLLHZL2019,JLZCLHZHL2021} without human annotation. 
The features learned by these models are agnostic to the tasks being evaluated, and recent literature has shown that they are well suited for downstream tasks such as phone recognition, speech recognition, and emotion recognition \cite{SBCA2019,9688093,9814838,9688253,li2023exploration}.
One of the most appealing strengths of these features is their linear separability, which allows easy extraction of desired information using a simple linear layer \cite{mohamed2022self,yang21c_interspeech}.

Despite the success of self-supervised speech representation learning (S3RL) models in evaluating diverse datasets with large vocabularies, most research has largely neglected to investigate their effectiveness on biased datasets such as in keyword spotting domain.
Industrial-scale keyword data often exhibits significant bias towards utterances that include designated keywords.
In particular, contrastive self-supervised learning \cite{VLV2018, BZMA2020} may be limited by a lack of diversity in the training data, causing the model to encode spurious noise to improve contrast. 
Such noise is not a desirable feature and may result in overfitting towards the model training.

Besides, the most previous S3RL research focuses on improving state-of-the-art performance on public benchmarks such as SUPERB \cite{yang21c_interspeech, lin2023utility} or internal proprietary data. 
However, these models are typically large and require significant computational complexity and storage \cite{BZMA2020, hsu2021hubert}.
The 12-layer transformer-based self-supervised models, with 95 million parameters, are particularly challenging to deploy in on-device keyword spotting.
End devices have extremely limited computational and storage resources, making it infeasible to use such large models for budget-constrained real-time applications.

\begin{figure}[t]
  \centering
  \includegraphics[width=\linewidth]{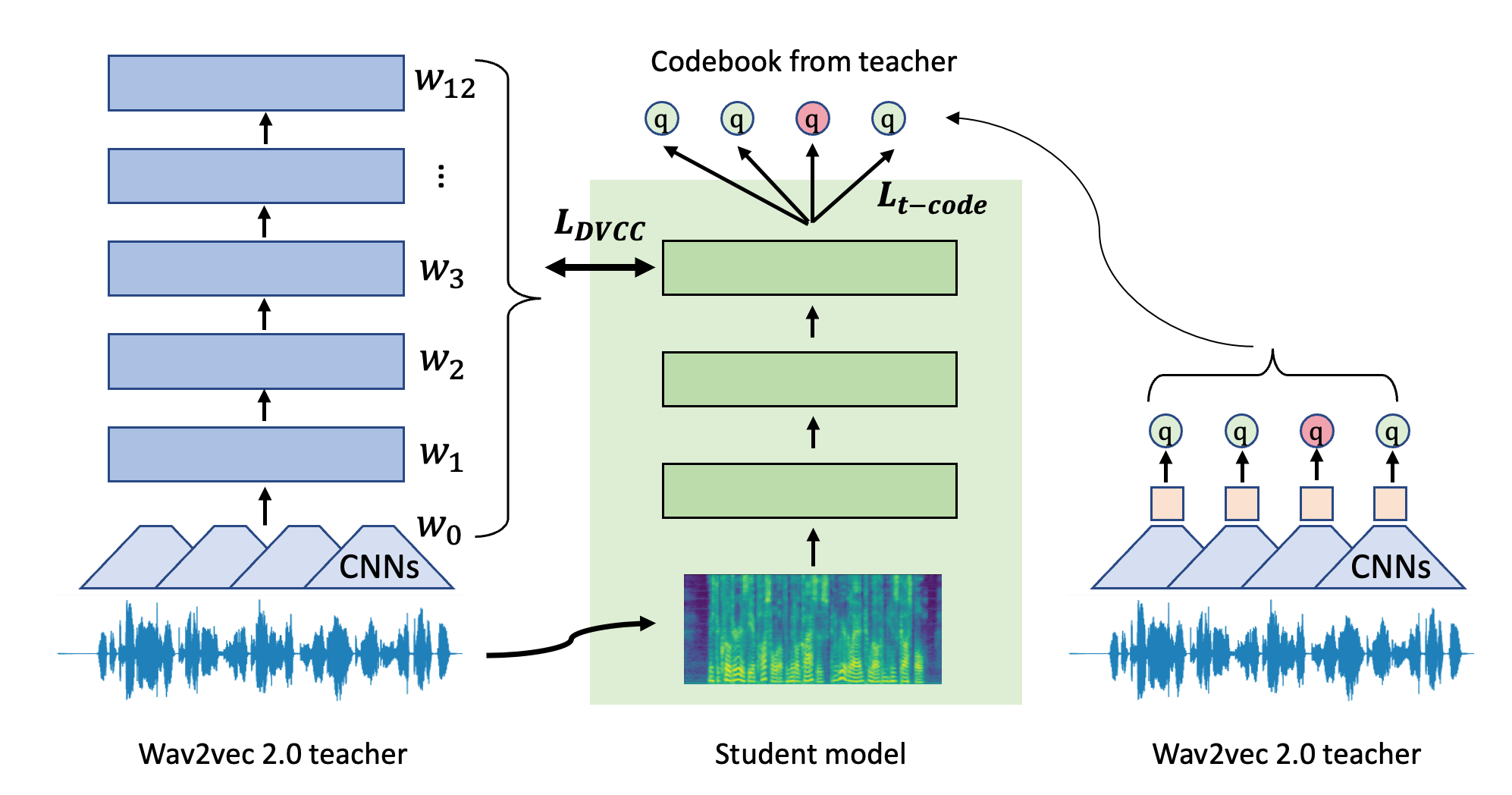}
  \vspace{-1.0em}
  \caption{Proposed knowledge distillation pipeline.}
  \vspace{-2.0em}
  \label{fig:overall}
\end{figure}

To overcome the challenges associated with data bias and model size on keyword spotting, we utilized knowledge distillation techniques in S3RL models \cite{hinton2015distilling}.
Our approach introduced two novel distillation techniques: dual-view cross-correlation distillation, and teacher codebook distillation.
To address the issue of model size, we distilled knowledge from large, self-supervised models (teacher) to smaller, on-device lightweight models (students) \cite{peng-etal-2021-shrinking,lin2022compressing,lee22p_interspeech,huang2023ensemble}. 
Specifically, we used the Wav2vec 2.0 \cite{BZMA2020} trained with LibriSpeech 960 hour set as the teacher model and transferred the knowledge to a 3-layer transformer architecture with 21 million or 1.6 million parameters as the student model.
Compared to prior techniques such as DistilHuBERT \cite{chang2022distilhubert} and LightHuBERT \cite{wang22t_interspeech}, which employed distance-based distillation on single frames, our dual-view cross-correlation distillation method considers the interdependence between samples and feature dimensions. 
This strategy takes into consideration the correlation between each feature of an utterance and each dimension, thereby optimizing its effectiveness in the distillation procedure even when the training data lacks diverse and high-quality negative samples.

To further alleviate the in-domain data bias during pre-training the on-device model, we proposed leveraging the codebook in teacher models, which are trained on a more diverse dataset, to improve the distillation process using a smaller and biased dataset.
Specifically, the student model is trained with native wav2vec 2.0 objective, while the positive and negative samples are the quantized vectors drawn from teacher model with teacher codebook.
The teacher codebook can be viewed as a compact representation of large, diverse speech data.
By using this codebook, we addressed the missing information issue from our biased data and also alleviate the training of the codebook requiring additional diversity loss.

We conducted the experiments on a de-identified 16,600 hours in-house keyword spotting dataset. 
The result showed that the proposed knowledge distillation based S3RL model outperforms the baselines on both normal and noisy conditions.  
The ablation study demonstrated that the introduced dual-view cross-correlation regularization surpass both previously L1 and Cosine similarity methods and single-view approaches on distillation. 
We also observed that the distillation method that utilized the teacher codebook as the training objective produced better results than the method without integrating the teacher codebook, especially under noisy conditions.
The superior performance of the proposed model emphasized the efficiency and effectiveness of innovative knowledge distillation methods in developing on-device constrained self-supervised models for keyword spotting tasks.
Our contributions can be summarized as:
\begin{enumerate}
    \item Developed an efficient and effective on-device constrained self-supervised model for keyword spotting task through knowledge distillation method.
    \item Devised two novel knowledge distillation techniques, dual-view cross-correlation distillation and teacher codebook distillation, to facilitate the effectiveness and robustness of knowledge transfer.
    \item Conducted an extended analysis and ablation study to investigate the potential crucial factors and benefits of S3RL knowledge distillation on keyword spotting tasks.
\end{enumerate}

\vspace{-0.2em}
\section{Methodology}
\vspace{-0.2em}
This section provides an overview of our S3RL based knowledge distillation system, which encompasses both the general teacher-student architecture and our newly proposed distillation mechanism.
The overall system design can be observed in Figure 1.
Initially, we will outline the general framework for knowledge distillation, while a more comprehensive explanation of the proposed distillation design will be presented in section 2.1 and section 2.2.

Given an input utterance $X$, the self-supervised teacher model generates a sequence of hidden features $h_1, h_2, ..., h_T$, where $T$ is the number of time frame.
The same input utterance is then fed into the student model, which may incorporate an augmented or distorted view of the input, generating another set of hidden features $o_1, o_2, ..., o_T$. 
The objective of knowledge distillation is designed to encourage the student model learns high-fidelity representation of the teacher model.
We illustrate one potential method by employing the L1 distance and cosine distance as a metric of similarity for the two features, and the loss function is defined as follows:
\begin{equation}
    L = \sum_{t=1}^{T} \bigl[ \lVert h_t - o_t \rVert_1 - \lambda \sigma \left(\cos{(h_t, o_t)}\right) \bigr],
\end{equation}
where $\lambda$ controls the weighting and is set to 1 in \cite{chang2022distilhubert}.

Unlike previous approaches that used frame-wise features \cite{chang2022distilhubert, wang22t_interspeech}, we focus on distilling utterance-wise features to avoid capturing variations in individual frames \cite{shor22_interspeech}.
This results in overall utterance-wise representations that are more representative and effective for downstream keyword classification. The loss function can be adjusted as follow:

\begin{equation}
    L = \lVert \overline{h} - \overline{o} \rVert_1 - \lambda \sigma \left(\cos{(\overline{h}, \overline{o})}\right), 
\end{equation}
where $\overline{h}$ and $\overline{o}$ are the averaged features over time.
\vspace{-0.5em}
\subsection{Dual-View Cross-Correlation Distillation}
\vspace{-0.5em}
Inspired by the work of Barlow-Twins and its successors \cite{zbontar2021barlow, mehrotra2023resource,liu2022selfsupervised}, we propose a novel dual-view cross-correlation mechanism to facilitate contrastive knowledge distillation in our proposed teacher-student design. 
Specifically, our method regularizes two cross-correlation matrices on batch-view and feature-view to reduce feature dimensional redundancy and generalize contrast operation, respectively. 
In our modeling process, we define two sets of features: a batch of teacher features represented as $H \in \mathbb{R}^{b \times d}$ and a batch of student features represented as $O \in \mathbb{R}^{b \times d}$, where $b$ indicates the batch size and $d$ is the feature dimension.
We apply average pooling along the time axis for each utterance to obtain the feature shape of $[d]$ from $[T, d]$.
Notably, our approach calculates the correlation matrix between the batches of teacher and student features from both feature-view and batch-view.

As shown in in Figure 2, we refer the feature-view as redundancy reduction, which calculates the cross-correlation matrix $C$ as follow:
\begin{equation}
    C_{ij} = \frac{\sum_b H_{bi} O_{bj}}{\sqrt{\sum_b (H_{bi})^2}\sqrt{\sum_b (O_{bj})^2}},
\end{equation}
where $C$ is a square matrix of shape $\mathbb{R}^{d \times d}$.
The goal of $C$ is to match identity matrix, where the on-diagonal elements are 1 and the off-diagonals are 0.
The objective can be formulate as follow:
\begin{equation}
    L_C = \sum_i (C_{ii} - 1)^2 + \alpha \sum_{i, j \neq i} C_{ij}^2
\end{equation}
The goal of applying the dot product to the batch dimension is to minimize redundancy in each feature dimension and produce a more streamlined student feature. This process seeks to make the student feature as compact as possible.

Regarding the batch-view, we generalize the contrast operation on the feature dimension, which is formulated as:
\begin{equation}
    G_{ij} = \frac{\sum_d H_{id} O_{jd}}{\sqrt{\sum_d (H_{id})^2}\sqrt{\sum_d (O_{jd})^2}},
\end{equation}
where $G$ is of shape $\mathbb{R}^{b \times b}$.
The objective for $G$ is:
\begin{equation}
    L_G = \sum_i (G_{ii} - 1)^2 + \beta \sum_{i, j \neq i} G_{ij}^2
\end{equation}
It aims to maximize the similarity between features from the same sample and minimize the correlation between features from different samples. By combining the two views, we get:
\begin{equation}
    L_{\text{DVCC}} = L_C / sg(L_C) + L_G / sg(L_G),
\end{equation}
The notation $sg$ is used to denote stop gradient with both terms scaled to 1.
The complete loss function dynamically integrates both components, thus eliminating the need to manually adjust the weights of the two terms. This approach saves effort and rationalizes the optimization process.

\begin{figure}[t]
  \centering
  \includegraphics[width=\linewidth]{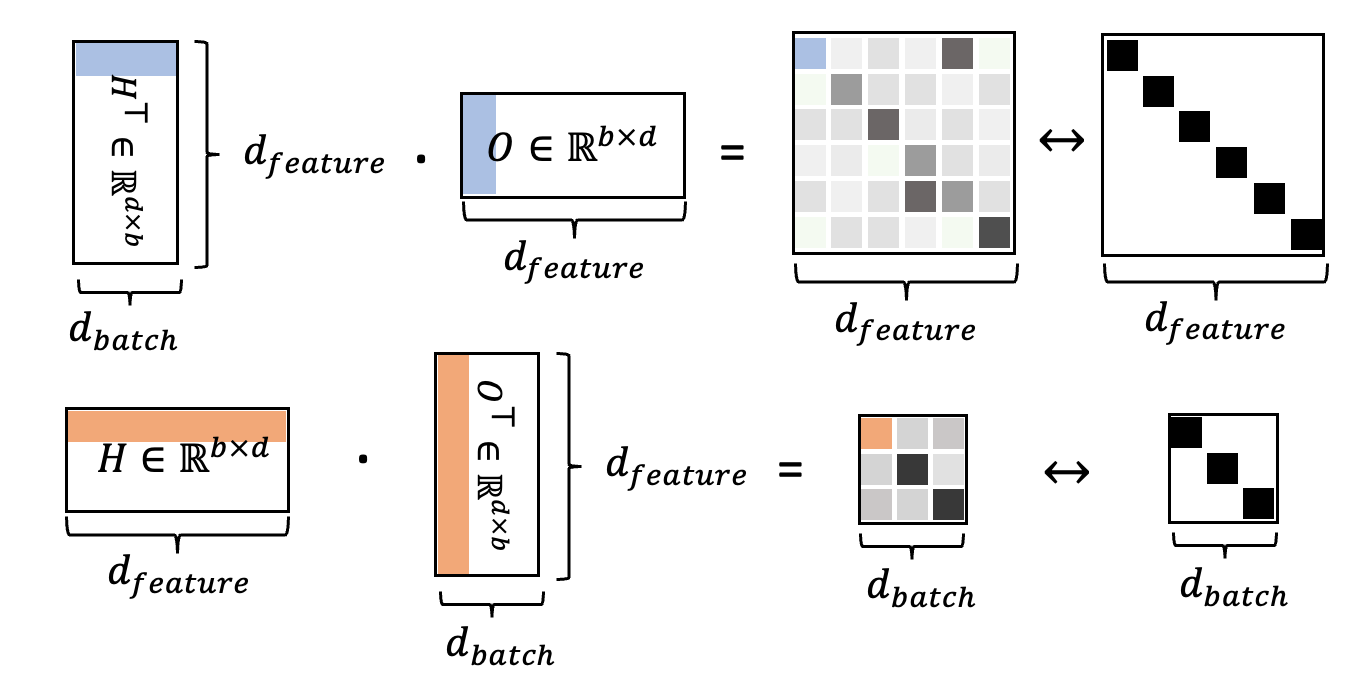}
  \vspace{-2em}
  \caption{Distillation using dual-view cross-correlation, where the top section illustrates the feature-view and the bottom section illustrates the batch-view. The figure depicts features with a dimension of 6 and a batch size of 3.}
  \label{fig:dual_view}
  \vspace{-2em}
\end{figure}

\vspace{-0.5em}
\subsection{Teacher Codebook Distillation}
\vspace{-0.5em}
When pre-training the student model on an in-domain biased dataset, there is a lack of diversity in the trained codebook, which presents a significant challenge for selecting sample pairs and representing unseen entries, especially in the case of the proposed utterance-wise representation on keyword spotting. 
To cope with this issue, we use a more robust codebook from the teacher model that has been trained on a more diverse speech dataset and distill the teacher codebook knowledge into the student model during pre-training.
We specifically employ the pre-trained Wav2vec 2.0 model as the teacher, which has been trained on the LibriSpeech 960 hour set. 
We train the student model with the same Wav2vec 2.0 objective as the teacher and select both positive and negative samples from the quantized features obtained from the teacher model during the mask prediction phase. 
This approach prevents the student from learning a codebook that captures subtle noise and effectively boosts the distillation computation by using only 5\% of the teacher’s total parameters, specifically the CNN layers and the codebook.
The teacher-student robust contrastive loss is defined as follows:
\begin{equation}
    L_{\text{t-code}} = - \sum_t \log \frac{exp(cos(o_t, k_t))}{\sum_{\Tilde{k} \sim K_t} exp(cos(o_t, \Tilde{k}))},
\end{equation}
where $o_t$ denotes prediction made by the student model, $k_t$ refers to positive quantized codebook from the teacher model, and $K_t$ represents the set of one positive $k_t$ and $N$ negative samples sampled from the teacher codebook.
\vspace{-0.5em}
\subsection{Combined Distillation Objective}
\vspace{-0.5em}
In order to enhance the performance of knowledge distillation in the S3RL model, we integrate the proposed dual-view cross-correlation and robust codebook distillation mechanisms.
By leveraging the unique advantages of both approaches, we can create a more effective distillation framework.
To execute our approach, we adopt a combined objective for the student model, as follows:
\begin{equation}
L_{\text{combined}} = L_{\text{DVCC}} + \gamma L_{\text{t-code}},
\end{equation}
The hyperparameter $\gamma$ is utilized to regulate the balance between the two objectives and can be adjusted to optimize the overall performance of the student model..

\section{Dataset and Implementation}
\subsection{Dataset}
\vspace{-0.5em}
We collected 16,600 hours of de-identified audio recordings in various front-end conditions for the Alexa keyword detection task. All data was processed into 64-D LFBE spectrograms using an analysis window of 25ms and a shift-size of 10ms. The dataset was split into 85 hours for validation, 85 hours for testing, and the remaining hours for training. To evaluate the model robustness, the test set was further divided into the normal condition with clean speech and playback condition with increased noise. The keyword labels were based on human annotation and underwent a quality check inspection.

\subsection{Implementation}
\vspace{-0.5em}
We used wav2vec 2.0 as the teacher model, comprising 7 CNN layers and 12 transformer layers, with a total of 95 million parameters. To meet device budget constraints for student model, we removed the CNN layers, reducing computation by approximately 33\% \cite{lin2022melhubert}. The student model directly takes LFBE features as input and consists of 3 transformer layers with a hidden size of 768 and 256, resulting in 21 million parameters with a 78\% size reduction, and 1.6 million parameters with a 98\% size reduction. We trained the student model with Adam optimizer for 15 epochs, each epoch comprising 5,000 update steps with a batch size of 512. For distillation, we used a learned weighted sum of all the hidden layers of the teacher model as the teacher feature. For fine-tuning, we added a linear layer on the last transformer layer of the student model and used cross-entropy loss to modify model parameters. The student model was fine-tuned for 30 epochs, each epoch having 5,000 steps with a batch size of 2,048. We set $\alpha$ and $\beta$ to $5 \times 10^{-3}$, and $\gamma$ to $1$. Same dataset was used for both knowledge distillation and downstream training phases, which was designated for keyword detection.

\vspace{-0.5em}
\subsection{Evaluation Metrics}
\vspace{-0.5em}
We evaluated the performance of our method on our internal dataset by measuring the false acceptance rate (FAR) at a fixed false rejection rate (FRR) in comparison to the baseline model. The FRR is the proportion of false negatives to true positives for a given keyword at the operating point (OP) of the baseline model. We determined the OP at which our proposed approach exhibited a comparable FRR and employed that same OP to calculate the corresponding FAR, which represents the ratio of false positives to true negatives.

\section{Results}

\subsection{Baseline Comparison}
\vspace{-0.5em}
To evaluate the impact of knowledge distillation on lightweight keyword spotting in self-supervised speech representation learning, we established a baseline model without knowledge distillation.
We pre-trained and fine-tuned the baseline model using the student architecture as the backbone. 
The outcomes in Table \ref{tab:result_1} demonstrated that the proposed dual-view cross-correlation based knowledge distillation approach outperforms the baseline by 14.6\% and 21.3\% with respect to the relative false acceptance rate (FAR) in normal and playback conditions, respectively.
Additionally, we replicated the DistilHuBERT method by adjusting the objectives to operate on utterance-wise features and predicting the weighted sum of the features from all teacher layers instead of making three predictions into respective layers.
The result revealed that the dual-view approach led to an improvement of more than 8\% relative FAR on our in-house alexa test set under both normal and playback (noisy) conditions.
Table \ref{tab:result_1} also displayed the outcomes of integrating the teacher-codebook into the dual-view cross-correlation process during knowledge distillation. 
The results demonstrated a consistent improvement in the relative FAR of the combined\_large model (21M parameter model), yielding values of 0.850 and 0.762 in normal and playback conditions, respectively. 
Subsequently, we further reduced the student model size to 1.6M parameters.
The findings indicated that this ultra-lightweight student model outperforms the baseline model with same model size by 10\%, and achieved comparable relative FAR compared with the 21M parameters baseline.
These results substantiate the efficiency and effectiveness of knowledge distillation-based S3RL for keyword spotting tasks subject to on-device budget constraints.

\begin{table}[!htb]
\begin{threeparttable}
  \caption{Experiment results on Alexa keyword spotting}
  \label{tab:result_1}
  \centering
  \begin{tabular}{lcccc}
    \toprule
     \multirow{2}{*}{\shortstack[c]{\textbf{Method}}} & \multirow{2}{*}{\shortstack[c]{\textbf{Model Size}}} & \multicolumn{2}{c}{\textbf{Relative FAR}} \\
     \cline{3-4}
     & & \textbf{Normal} & \textbf{Playback} \\
    \midrule
    Baseline w/o KD & 21M & 1.0 & 1.0  \\
    Ultra-light w/o KD & 1.6M & 1.17 & 1.22  \\
    DistilHuBERT & 21M & 0.937 & 0.901    \\
    Feature-view & 21M & 0.853 & 0.817         \\
    Batch-view & 21M & 0.861 & 0.813    \\
    Dual-view & 21M & 0.854 & 0.787   \\
    \midrule
    w/o T-codebook & 21M & 0.907 & 0.884  \\
    w/ T-codebook & 21M & 0.903 & 0.841           \\
    \midrule
    Combined\_large & 21M & 0.850 & 0.762\\
    Combined\_small & 1.6M & 1.07 & 1.09\\
    \bottomrule
  \end{tabular}
  \begin{tablenotes}
  \item\footnotesize \textbf{Legend:} \textbf{Relative FAR}=relative false acceptance rate compared to baseline at fixed false rejection rate; \textbf{T-codebook}=teacher codebook; \textbf{Dual-view}=Batch-view + Feature-view; \textbf{Combined\_large/small}= Dual-view + T-codebook. 
  \end{tablenotes}
  \vspace{-1em}
\end{threeparttable}
\end{table}

\vspace{-1.0em}

\subsection{Single-View vs Dual-View}
To further examine the advantages of the proposed dual-view cross-correlation approach, we conducted an ablation study to compare the performance of single-view (batch-view or feature-view) and dual-view distillation.
The results presented in Table \ref{tab:result_1} demonstrated that, under normal testing condition, batch-view, feature-view, and dual-view distillation produced similar results, with feature-view distillation exhibiting slightly better performance than the other two approaches.
This demonstrated that the reduction of redundancy among each feature dimension in the feature-view facilitates the generalization of the model.
In contrast, when subjected to noise, the dual-view distillation method surpassed the other approaches by 2.6\% and 3\% relative FAR, which indicated that the contrastive nature between different samples in the batch-view is an essential element for learning a more robust model.
\vspace{-0.5em}
\subsection{Teacher Codebook vs Codebook from Scratch}
\vspace{-0.5em}
By comparing the results presented in the second group of rows in Table \ref{tab:result_1}, we noticed that the distillation approach that employed the teacher codebook as the training objective performed better than the approach without integrating the teacher codebook during training, particularly in noisy conditions.
The model obtained even better results by combining the dual-view distillation with the teacher codebook.
We also observed that the benefits of the teacher codebook were less prominent when combined with dual-view distillation under normal conditions, but still resulted in a 2.5\% relative (FAR) gain under playback conditions.
These findings confirmed our hypothesis that the teacher codebook serves as a more diverse speech representation, thereby augmenting the effectiveness of contrastive self-supervised learning.
\vspace{-0.5em}
\subsection{Layer Selection from Teacher Model}
\vspace{-0.5em}
While developing the model, we discovered that by training only a linear classifier on keyword spotting, the features from layers 5, 6, 7, and 8 of the teacher model outperform those from other layers on a linear classifier, with the final layer exhibiting the weakest performance \cite{9688093,pasad2022comparative}. 
Consequently, we conducted additional experiments to assess the effectiveness of distillation using only the selected layers, as opposed to employing information from all layers. 
In previous experiments, we trained 13 weights to aggregate the CNN output feature and all 12 transformer layer features from the teacher model.
As in sub-layer experiment, we first utilized four weights to compute a weighted sum of features from layers 5 to 8 for distillation. 
Moreover, we trained another sub-layer model to nullify the information from layers 5 to 8 and only distill information from the remaining layers. 
We evaluated different layer distillation techniques using dual-view cross-correlation distillation.
\vspace{-0.5em}
\begin{table}[!htb]
  \caption{Dual-View Cross-Correlation (DVCC) distillation from selected teacher layers. }
  \label{tab:result_layer}
  \centering
  \begin{tabular}{lccc}
    \toprule
    \multirow{2}{*}{\shortstack[c]{\textbf{Method}}} & \multicolumn{2}{c}{\textbf{Relative FAR}} \\
     \cline{2-3}
     & \textbf{Normal} & \textbf{Playback} \\
    \midrule
    Baseline w/o KD & 1.0 & 1.0  \\
    DVCC w/ $\text{Layer}_{0-12}$ & 0.854 & 0.787 \\
    DVCC w/ $\text{Layer}_{5-8}$ & 0.806 & 0.713\\
    DVCC w/ $\text{Layer}_{0-4, 9-12}$ & 0.855 & 0.790\\
    \bottomrule
  \end{tabular}
  \vspace{-1em}
\end{table}
\vspace{-1.0em}
\begin{figure}[!htb]
  \centering
  \includegraphics[width=\linewidth]{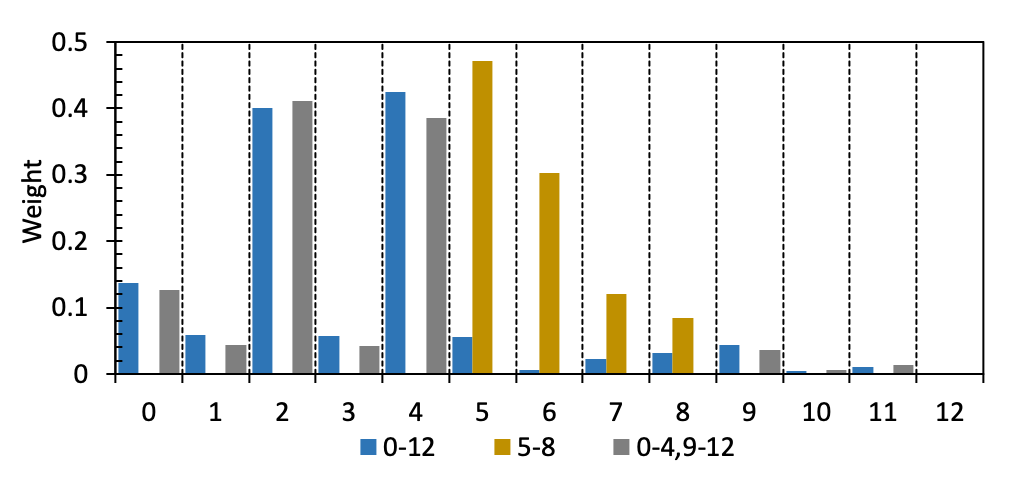}
  \vspace{-2em}
  \caption{Learned weighting for the teacher features with different sets of layers.}
  \label{fig:weight}
\end{figure}
\vspace{-1.0em}

Table \ref{tab:result_layer} illustrates that distilling information from layers 5 to 8 led to the most exceptional overall performance. 
We also noted that distilling information from the remaining layers produced results comparable to those of distilling from all layers.
Further analysis revealed that the most significant learned weights for distillation information from layers 0-12 and layers 0-4,9-12 were found in layers 0 (CNN output), 2, and 4, while the most significant learned weights for layers 5-8 were found in the respective layers as shown in Figure \ref{fig:weight}.
This indicates that the comparable performance attained by employing the remaining layers may be due to the fact that the model distilling all layers fails to entirely capture information from layers 5-8.
This finding indicates that a straightforward layer selection can lead to substantial benefits since the model is ignorant of downstream tasks during the distillation process.

\section{Conclusions and Future Work}
\vspace{-0.3em}
In this study, we proposed a new self-supervised speech representation learning (S3RL) architecture for on-device keyword spotting tasks that utilizes two novel knowledge distillation methods: dual-view cross-correlation and teacher codebook distillation. 
Our experiments on a biased keyword spotting dataset confirmed the effectiveness and robustness of our approach, highlighting its potential for improving the performance of S3RL knowledge distillation and providing promising avenues for research direction in this field.
To extend the applicability of the introduced approach, our future work will include experiments on other downstream tasks using different datasets, as this research has solely focused on on-device keyword spotting. 
This will enable us to evaluate the generalizability of our proposed method. 

\nocite{*}
\bibliographystyle{IEEEtran}
\bibliography{mybib}

\begin{thebibliography}{10}
\providecommand{\url}[1]{#1}
\csname url@samestyle\endcsname
\providecommand{\newblock}{\relax}
\providecommand{\bibinfo}[2]{#2}
\providecommand{\BIBentrySTDinterwordspacing}{\spaceskip=0pt\relax}
\providecommand{\BIBentryALTinterwordstretchfactor}{4}
\providecommand{\BIBentryALTinterwordspacing}{\spaceskip=\fontdimen2\font plus
\BIBentryALTinterwordstretchfactor\fontdimen3\font minus
  \fontdimen4\font\relax}
\providecommand{\BIBforeignlanguage}[2]{{%
\expandafter\ifx\csname l@#1\endcsname\relax
\typeout{** WARNING: IEEEtran.bst: No hyphenation pattern has been}%
\typeout{** loaded for the language `#1'. Using the pattern for}%
\typeout{** the default language instead.}%
\else
\language=\csname l@#1\endcsname
\fi
#2}}
\providecommand{\BIBdecl}{\relax}
\BIBdecl

\bibitem{mohamed2022self}
A.~Mohamed, H.-y. Lee, L.~Borgholt, J.~D. Havtorn, J.~Edin, C.~Igel,
  K.~Kirchhoff, S.-W. Li, K.~Livescu, L.~Maal{\o}e \emph{et~al.},
  ``Self-supervised speech representation learning: A review,'' \emph{IEEE
  Journal of Selected Topics in Signal Processing}, 2022.

\bibitem{yang21c_interspeech}
S.~wen Yang, P.-H. Chi, Y.-S. Chuang, C.-I.~J. Lai, K.~Lakhotia, Y.~Y. Lin,
  A.~T. Liu, J.~Shi, X.~Chang, G.-T. Lin, T.-H. Huang, W.-C. Tseng, K.~tik Lee,
  D.-R. Liu, Z.~Huang, S.~Dong, S.-W. Li, S.~Watanabe, A.~Mohamed, and
  H.~yi~Lee, ``{SUPERB: Speech Processing Universal PERformance Benchmark},''
  in \emph{Interspeech}, 2021.

\bibitem{CHTG2019}
Y.-A. Chung, W.-N. Hsu, H.~Tang, and J.~Glass, ``An unsupervised autoregressive
  model for speech representation learning,'' in \emph{Interspeech}, 2019.

\bibitem{ling2020decoar}
S.~Ling and Y.~Liu, ``Decoar 2.0: Deep contextualized acoustic representations
  with vector quantization,'' \emph{arXiv preprint arXiv:2012.06659}, 2020.

\bibitem{yang2022}
G.-P. Yang, S.-L. Yeh, Y.-A. Chung, J.~Glass, and H.~Tang, ``Autoregressive
  predictive coding: A comprehensive study,'' \emph{IEEE Journal of Selected
  Topics in Signal Processing}, 2022.

\bibitem{VLV2018}
A.~van~den Oord, Y.~Li, and O.~Vinyals, ``Representation learning with
  contrastive predictive coding,'' \emph{arXiv:1807.03748}, 2018.

\bibitem{BZMA2020}
A.~Baevski, Y.~Zhou, A.~Mohamed, and M.~Auli, ``wav2vec 2.0: A framework for
  self-supervised learning of speech representations,'' in \emph{NeurIPS},
  2020.

\bibitem{9053541}
W.~Wang, Q.~Tang, and K.~Livescu, ``Unsupervised pre-training of bidirectional
  speech encoders via masked reconstruction,'' in \emph{ICASSP}, 2020.

\bibitem{JLLLHZL2019}
D.~Jiang, X.~Lei, W.~Li, N.~Luo, Y.~Hu, W.~Zou, and X.~Li, ``Improving
  transformer-based speech recognition using unsupervised pre-training,''
  \emph{arXiv:1910.09932}, 2019.

\bibitem{JLZCLHZHL2021}
D.~Jiang, W.~Li, R.~Zhang, M.~Cao, N.~Luo, Y.~Han, W.~Zou, K.~Han, and X.~Li,
  ``A further study of unsupervised pretraining for transformer based speech
  recognition,'' in \emph{ICASSP}, 2021.

\bibitem{SBCA2019}
S.~Schneider, R.~C. Alexei~Baevski, and M.~Auli, ``wav2vec: Unsupervised
  pre-training for speech recognition,'' in \emph{Interspeech}, 2019.

\bibitem{9688093}
A.~Pasad, J.-C. Chou, and K.~Livescu, ``Layer-wise analysis of a
  self-supervised speech representation model,'' in \emph{2021 IEEE Automatic
  Speech Recognition and Understanding Workshop (ASRU)}, 2021.

\bibitem{9814838}
S.~Chen, C.~Wang, Z.~Chen, Y.~Wu, S.~Liu, Z.~Chen, J.~Li, N.~Kanda,
  T.~Yoshioka, X.~Xiao, J.~Wu, L.~Zhou, S.~Ren, Y.~Qian, Y.~Qian, J.~Wu,
  M.~Zeng, X.~Yu, and F.~Wei, ``Wavlm: Large-scale self-supervised pre-training
  for full stack speech processing,'' \emph{IEEE Journal of Selected Topics in
  Signal Processing}, 2022.

\bibitem{9688253}
Y.-A. Chung, Y.~Zhang, W.~Han, C.-C. Chiu, J.~Qin, R.~Pang, and Y.~Wu,
  ``w2v-bert: Combining contrastive learning and masked language modeling for
  self-supervised speech pre-training,'' in \emph{ASRU}, 2021.

\bibitem{li2023exploration}
Y.~Li, Y.~Mohamied, P.~Bell, and C.~Lai, ``Exploration of a self-supervised
  speech model: A study on emotional corpora,'' in \emph{SLT}, 2023.

\bibitem{lin2023utility}
G.-T. Lin, C.-L. Feng, W.-P. Huang, Y.~Tseng, T.-H. Lin, C.-A. Li, H.-y. Lee,
  and N.~G. Ward, ``On the utility of self-supervised models for
  prosody-related tasks,'' in \emph{2022 IEEE Spoken Language Technology
  Workshop (SLT)}.\hskip 1em plus 0.5em minus 0.4em\relax IEEE, 2023.

\bibitem{hsu2021hubert}
W.-N. Hsu, B.~Bolte, Y.-H.~H. Tsai, K.~Lakhotia, R.~Salakhutdinov, and
  A.~Mohamed, ``Hubert: Self-supervised speech representation learning by
  masked prediction of hidden units,'' \emph{IEEE/ACM Transactions on Audio,
  Speech, and Language Processing}, 2021.

\bibitem{hinton2015distilling}
G.~Hinton, O.~Vinyals, and J.~Dean, ``Distilling the knowledge in a neural
  network,'' \emph{arXiv preprint arXiv:1503.02531}, 2015.

\bibitem{peng-etal-2021-shrinking}
Z.~Peng, A.~Budhkar, I.~Tuil, J.~Levy, P.~Sobhani, R.~Cohen, and J.~Nassour,
  ``Shrinking bigfoot: Reducing wav2vec 2.0 footprint,'' in
  \emph{sustainlp}.\hskip 1em plus 0.5em minus 0.4em\relax ACL, 2021.

\bibitem{lin2022compressing}
T.-Q. Lin, T.-H. Yang, C.-Y. Chang, K.-M. Chen, T.-h. Feng, H.-y. Lee, and
  H.~Tang, ``Compressing transformer-based self-supervised models for speech
  processing,'' \emph{arXiv preprint arXiv:2211.09949}, 2022.

\bibitem{lee22p_interspeech}
Y.~Lee, K.~Jang, J.~Goo, Y.~Jung, and H.~R. Kim, ``{FitHuBERT: Going Thinner
  and Deeper for Knowledge Distillation of Speech Self-Supervised Models},'' in
  \emph{Interspeech}, 2022.

\bibitem{huang2023ensemble}
K.-P. Huang, T.-h. Feng, Y.-K. Fu, T.-Y. Hsu, P.-C. Yen, W.-C. Tseng, K.-W.
  Chang, and H.-y. Lee, ``Ensemble knowledge distillation of self-supervised
  speech models,'' \emph{arXiv preprint arXiv:2302.12757}, 2023.

\bibitem{chang2022distilhubert}
H.-J. Chang, S.-w. Yang, and H.-y. Lee, ``Distilhubert: Speech representation
  learning by layer-wise distillation of hidden-unit bert,'' in
  \emph{ICASSP}.\hskip 1em plus 0.5em minus 0.4em\relax IEEE, 2022.

\bibitem{wang22t_interspeech}
R.~Wang, Q.~Bai, J.~Ao, L.~Zhou, Z.~Xiong, Z.~Wei, Y.~Zhang, T.~Ko, and H.~Li,
  ``{LightHuBERT: Lightweight and Configurable Speech Representation Learning
  with Once-for-All Hidden-Unit BERT},'' in \emph{Proc. Interspeech 2022},
  2022.

\bibitem{shor22_interspeech}
J.~Shor and S.~Venugopalan, ``{TRILLsson: Distilled Universal Paralinguistic
  Speech Representations},'' in \emph{Proc. Interspeech 2022}, 2022.

\bibitem{zbontar2021barlow}
J.~Zbontar, L.~Jing, I.~Misra, Y.~LeCun, and S.~Deny, ``Barlow twins:
  Self-supervised learning via redundancy reduction,'' in \emph{International
  Conference on Machine Learning}.\hskip 1em plus 0.5em minus 0.4em\relax PMLR,
  2021.

\bibitem{mehrotra2023resource}
\BIBentryALTinterwordspacing
A.~Mehrotra, A.~G. C.~P. Ramos, N.~D. Lane, and S.~Bhattacharya, ``Resource
  efficient self-supervised learning for speech recognition,'' 2023. [Online].
  Available: \url{https://openreview.net/forum?id=L9pW5fknjO}
\BIBentrySTDinterwordspacing

\bibitem{liu2022selfsupervised}
\BIBentryALTinterwordspacing
X.~Liu, Z.~Wang, Y.-L. Li, and S.~Wang, ``Self-supervised learning via maximum
  entropy coding,'' in \emph{Advances in Neural Information Processing
  Systems}, A.~H. Oh, A.~Agarwal, D.~Belgrave, and K.~Cho, Eds., 2022.
  [Online]. Available: \url{https://openreview.net/forum?id=nJt27NQffr}
\BIBentrySTDinterwordspacing

\bibitem{lin2022melhubert}
T.-Q. Lin, H.-y. Lee, and H.~Tang, ``Melhubert: A simplified hubert on mel
  spectrogram,'' \emph{arXiv preprint arXiv:2211.09944}, 2022.

\bibitem{pasad2022comparative}
A.~Pasad, B.~Shi, and K.~Livescu, ``Comparative layer-wise analysis of
  self-supervised speech models,'' \emph{arXiv preprint arXiv:2211.03929},
  2022.

\end{thebibliography}

\end{document}